\documentclass[journal,transmag]{IEEEtran}

\usepackage[pdftex]{graphicx}
\ifCLASSINFOpdf
\else
\fi
\usepackage{amsmath}
\usepackage{algorithm, algorithmic}

\usepackage{array}

\usepackage{stfloats}
\usepackage{pgfplots}
\usepackage{tikz}
\usepackage{multirow}
\usepackage{multicol}
\usepackage{amsfonts}

\begin{document}

%
\title{Partial Policy-based Reinforcement Learning for Anatomical Landmark Localization in 3D Medical Images}


\author{\IEEEauthorblockN{Walid Abdullah Al,
Il Dong Yun* (\textit{IEEE Member})}
\IEEEauthorblockA{Department of Computer and Electronic Systems Engineering,\\
Hankuk University of Foreign Studies, Yongin, South Korea}

\thanks{
*Email: yun@hufs.ac.kr}}

\markboth{}%
{Shell \MakeLowercase{\textit{et al.}}: Bare Demo of IEEEtran.cls for IEEE Transactions on Magnetics Journals}
%


\maketitle

\begin{abstract}
Utilizing the idea of long-term cumulative return, reinforcement learning (RL) has shown remarkable performance in various fields. We propose a formulation of the landmark localization in 3D medical images as a reinforcement learning problem. Whereas value-based methods have been widely used to solve RL-based localization problems, we adopt an actor-critic based direct policy search method framed in a temporal difference learning approach. In RL problems with large state and/or action spaces, learning the optimal behavior is challenging and requires many trials. To improve the learning, we introduce a partial policy-based reinforcement learning to enable solving the large problem of localization by learning the optimal policy on smaller partial domains. Independent actors efficiently learn the corresponding partial policies, each utilizing their own independent critic. The proposed policy reconstruction from the partial policies ensures a robust and efficient localization utilizing the sub-agents solving simple binary decision problems in their corresponding partial action spaces. Experiments with three different localization problems in 3D CT and MR images showed that the proposed reinforcement learning requires a significantly smaller number of trials to learn the optimal behavior compared to the original behavior learning scheme in RL. It also ensures a satisfactory performance when trained on a fewer images.    
\end{abstract}




%
\IEEEpeerreviewmaketitle

\section{Introduction}
%
%
%
%
\IEEEPARstart{L}{andmark} localization plays a vital role in medical image analysis, facilitating the automatic process for registration, classification, and segmentation \cite{sotiras2013deformable,han2017systems}. Besides speeding up the interpretation, it contributes to visualization and assessment-based applications. However, accurate landmark localization in 3D medical images is a challenging problem because of high inter-patient variations in terms of size, shape, and orientation, as well as the variations and artifacts caused by different parameter settings. Machine learning approaches are becoming more and more common to solve the localization problem under such variation. Standard approaches suggest classification or regression-based model in order to localize the landmarks. However, all of the previous learning approaches are mainly exploitative and may behave inconsistently for an exceptional test data. Long-term reward-oriented reinforcement learning (RL) algorithms offer ways to balance between exploration and exploitation, yielding a noteworthy performance in various fields of image processing \cite{mnih2015human,zhu2017target, mnih2013playing}. With a few instances of implementation in object localization in terms of a bounding box, RL-enforced landmark localization is rarely found.\par

Value function approximation (e.g., deep Q-Network (DQN) \cite{mnih2015human}) is a widely used method to solve the RL problem for large state and/or action spaces, suggesting an indirect behavior learning. Compared to such value-based methods, explicit behavior learning by directly approximating the policy function \cite{sutton2000policy} has the advantage of a better convergence. However, direct policy search method suffers from the high variance problem \cite{konda2000actor}. Actor-critic RL performs direct policy approximation while utilizing an additional value function approximator to reduce the variance, thus taking advantage of both the policy and value-based methods \cite{grondman2012survey}. Nevertheless, a good exploration in order to obtain the optimal policy in a large space is challenging. Despite remarkable propositions and improvement to maintain the balance between exploration and exploitation, RL practically faces problem to successfully learn a task in a large state and/or action space and requires many trials \cite{duan2016rl}.\par

\begin{figure}[h]
\centering
\includegraphics[width=\columnwidth]{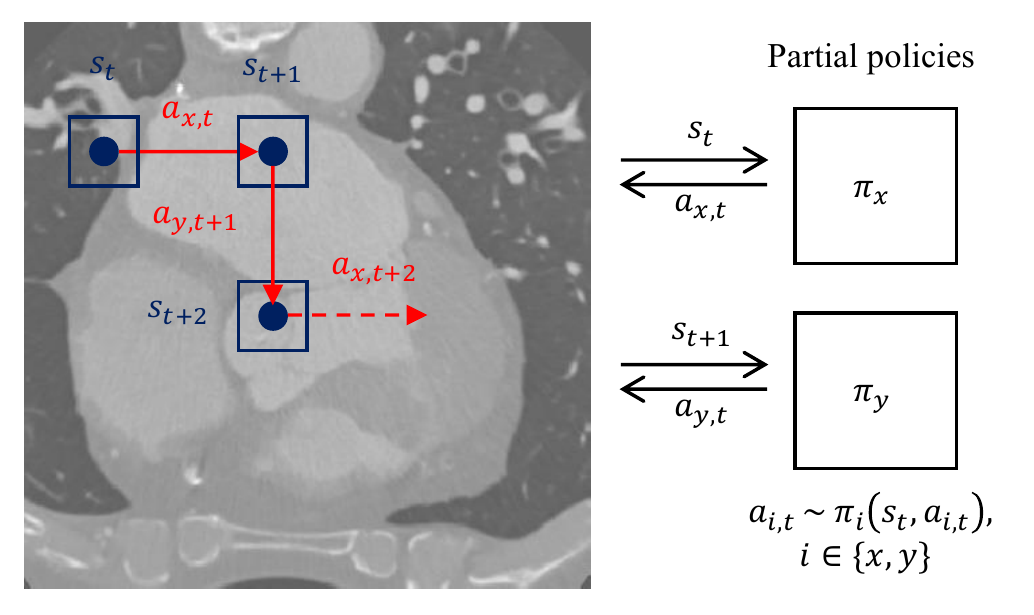}
\caption{{\bf Landmark localization process using the partial policies.} A 2D case is used for illustration purposes. An agent initiated at any position at time $t$, observes the corresponding state $s_t$ and decides an action to move towards the target landmark. Instead of a single RL agent with a single policy $\pi$, multiple sub-agents with simpler partial policies $\pi_x$ and $\pi_y$ repeatedly contribute to the successive state transitions. $a_{i,t}$ refers to the partial action at $t$, sampled from the partial policy distribution $\pi_i$ on the $i$-th sub-action space, for the current state $s_t$.}
\label{locintro}
\end{figure}

In this paper, we formulate the landmark localization problem as a sequential decision-making problem in RL, where an agent initiated at a random position inside a 3D medical image (i.e., volume) observes the current state and takes subsequent actions to move towards the target landmark. We suggest learning the policy function directly using an actor-critic approach because of its advantage over pure policy or value-based approach. To ensure a successful behavior learning within a significantly smaller number of trials, we introduce a partial policy-based reinforcement learning model where multiple sub-agents learn assigned micro-tasks to successfully learn the original task. Partial policies with respect to the micro-tasks are obtained by projecting the original policy onto smaller sub-action spaces, enabling a disintegration of the complex decision problem into a set of simpler problems. We allow independent actors to update the corresponding partial policy functions each utilizing its own value function (i.e., critic). Fig.~\ref{locintro} shows a schematic illustration of the localization process using the partial policies for a 2D case. \par

\section{Related Work}
Prior research on landmark localization in both medical image and computer vision concentrated on the model-based methods where a major focus was on the classification-based approaches. For example, Shotton et al. \cite{shotton2013real} utilized a trained forest to perform per-pixel classification to recognize body parts as an intermediary step required for localizing body-joints in depth images. Zheng et al. \cite{zheng} proposed a marginal space learning approach to localize aortic valve landmarks hierarchically, where a rough location is first obtained from a detected global object consisting of the landmarks, followed by a refinement session using local boosting tree based classifier. A generalized Procrustes analysis (GPA) was used to find the optimal global object. However, GPA does not guarantee the convergence of means \cite{procrust1}. Ionasec et al. \cite{ldmodel1} utilized a similar boosting tree classifier to localize the aortic valve landmarks. A forest classifier based method similar to \cite{shotton2013real} was also proposed for hand-joint localization in X-ray images. Nevertheless, classification approaches suffer from dataset imbalance problem because of the negligible positive samples, resulting in a biased classifier.\par


Regression-based approaches are becoming more and more popular instead of formulating the localization task as a classification problem. Regression models also showed significant improvement comparing to the classification models. These models suggest exploiting the predictions in different regions at a different distance from the target landmark. Criminisi et al. \cite{criminisi} proposed an efficient method for anatomy localization using a regression forest. Jung et al. \cite{rfw} proposed the random tree walk (RTW) method for localizing human body joints in depth images, where a regression tree is trained to estimate the direction to the target joint for each position. A walker then starts walking using the learned direction and the expectation of the stepped positions is considered as the resultant joint position. An implication of RTW for localizing the aortic valve landmarks can be found in our previous work \cite{cwplos}, where we performed a colonial walk initiating multiple random walks from different initial positions. The successful walker from the colony was elected by the minimum walk variance measure. Some joint models combining both the regression and classification approach also exist \cite{gall2011hough,schulter2014accurate}. Most recently, a stratified method is introduced by Oktay et al. \cite{stratified}, where image-patch driven local information as well as the global information in terms of organ-size, shape etc. are used for training.\par

There are a few approaches that do not fall under either of the above mentioned approaches. Simple connected component analysis \cite{ostia1}and coronary centerline tracking \cite{ostia2} is used for coronary ostia detection after aorta segmentation. However, robustness is challenged under image noise in case of connected component analysis, while centerline tracking algorithm has a high computational cost when operated for the whole surface of aorta. Elattar et al. \cite{mustafaself} detected coronary ostia and aortic hinges on the aortic root surface, which is segmented using connected component analysis.\par

Most of the previous learning-based approaches are generally exploitative and may face generalization problem. RL provides an explorative learning scheme, which has shown remarkable improvements in various fields of image processing. To the best of our knowledge, only one instance of RL-based landmark localization exists in the literature, whereas a few instances of bounding box-based object localization in natural images can be found. Caicedo et al. \cite{caicedo2015active} presented the object localization problem as a sequential decision-making problem in RL, where the process starts with a bounding-box covering the whole image, gradually applying transformation actions to the bounding box to finally localize the object. A better intersection-over-union (IoU) between the transformed box and the GT bounding box yields a positive reinforcement, and negative otherwise. The state is related to the image or sub-image inside the bounding box. Jie et al. \cite{jie2016tree} proposed a tree-structured RL with similar state and action representation. At each state, the agent applies two different actions i.e., scaling and translation, yielding two resultant states. Thus, the agent follows a recursive approach to finally find the object, representing a binary tree-like search. Ghesu et al. \cite{ghesu2016artificial} present the only RL implementation in anatomical landmark localization in medical images, where the agent makes sequential position update actions to reach the landmark. The state is defined to be the region-of-interest (ROI) around the corresponding position. Relative distance change is used as the rewarding scheme. All these approaches mainly implemented Q-learning \cite{mnih2014recurrent} to learn the action-value function while learning the optimal behavior indirectly. Direct behavior learning through optimizing the policy function shows better convergence comparatively, while having the problem of high variance. The actor-critic approach also approximates the policy function, however, uses an additional value function as the critic to reduce the variance \cite{grondman2012survey}.\par

Our work focuses on RL formulation for landmark localization in 3D medical images, where the agent action follows a definition similar to \cite{ghesu2016artificial}. Unlike the previous approaches, we directly approximate the policy function following the actor-critic approach, where a state-value function is used as the policy evaluator or the critic. Moreover, we propose to learn multiple partial policies on different sub-action spaces instead of a single complex policy on the original action space, in order to improve the slow learning problem of RL and ensure a more robust localization.\par

The rest of the paper is organized as follows. Section~\ref{sec3} describes the formulation of landmark localization for the actor-critic RL. Section~\ref{sec4} presents the partial policy-based RL for localization. Section~\ref{sec5} reports the experimental evaluation of the proposed approach. Finally, Section~\ref{sec6} presents the concluding remarks.

\section{Landmark Localization as RL}
\label{sec3}
In our reinforcement learning-based localization scheme, an RL agent initiated at a random position interacts with the volume by taking consecutive discrete actions sampled from the learned policy distribution for the observed state at the current position, to finally reach the target landmark. During training, the agent tries to attain an optimal policy that maximizes the long-term return formulated as a discounted cumulative reward. The following is the description of the key elements of the Markov decision process (MDP) in the RL-wrapped localization scheme. \par

\subsection{State}
We represent the state as a function of the agent-position. For any position $\boldsymbol{q}$, the corresponding state $\boldsymbol{s}=\mathcal{S}(\boldsymbol{q})$ refers to a stack of the axial, coronal, and sagittal sub-images observed through a squared window centered at the corresponding position. Thus, we allow the agent at any position to observe an $m \times m \times 3$ block of surrounding voxels. Here, $m$ is the window size. Such state is useful to provide a pseudo-3D view but requires less storage in the experience replay memory. This also helps traverse the state through a usual 2D CNN (of the policy and value networks), treating it as a 3-channeled image.\par

\subsection{Action}
Similar to Ghesu et al.'s approach \cite{ghesu2016artificial}, a discrete action space is considered where agent can take a unit step along either of the axes to update its position to a neighbouring voxel. Therefore, the agent holds three degrees of freedom to move, enabling six actions i.e., $right, left, up, down, slice\_{}forward, slice\_{} backward$. The first four moves are along the axial slice (X and Y axes), whereas, the last two actions allow the agent to jump across the slices moving along Z-axis. We represent our action space as follows: 
\begin{equation}\label{action_space} 
\mathcal{A} = \{x^+, x^-, y^+, y^-, z^+, z^-\}
\end{equation}
where $x^+$ , $x^-$, $y^+$, $y^-$, $z^+$ and $z^-$ represent $right$, $left$, $up$, $down$, $slice\_{}forward$ and $slice\_{}backward$, respectively.\par

Using these simple actions yields a rather simple and deterministic transition. For a given position $\boldsymbol{q}$ and action $a$, the transitioned position $\boldsymbol{q'}$ can easily be obtained using the following transition function:
\begin{equation}\label{transition_model}
\begin{split}
& \boldsymbol{q'} = \mathcal{T}(\boldsymbol{q}, a)= \big( q_x + U_x(a), q_y + U_y(a), q_z + U_z(a) \big)\\
& U_i(a)=
\begin{cases}
\eta, & \text{if } a=i^+\\
-\eta, & \text{if } a=i^-\\
0, & \text{otherwise } 
\end{cases}
\\
& i \in \{x, y, z\}
\end{split}
\end{equation}
Here, $\eta$ is the length of a unit-step. $q_x, q_y, $ and $q_z$ are the components of $\boldsymbol{q}$ along different axes. We denote such transition as $(\boldsymbol{q},a,\boldsymbol{q'})$. There could be an additional action for no transition where agent remains at its current position without moving so that we can know that it has reached its destination landmark. However, adding such action would make the optimal policy finding harder because the state-action space (that should be explored) would become larger. Moreover, comparing to other actions, this action can render positive reward only for one state in the whole volume. Thus the model will suffered from sample selection bias and highly unlikely to trigger this action. Finally, we can ensure a satisfactory localization by using only just the aforementioned six actions. Because, eventually it would converge the target and move back and forth creating an oscillation of an amplitude of 1-2 voxels. The final localized position is the centroid of the oscillation, and may be approximated by taking the expectation of last few steps.\par

\subsection{Reward}
The agent at any position inside a 3D volume targets at choosing an action that maximizes the discounted cumulative reward. Therefore, we should encourage the agent to come closer to the target by giving an appropriate reward. We propose to use a simple binary reward function, where a positive reward is given if an action leads the agent closer to target landmark, and a negative reward is given otherwise. The reward is immediate after each action. The Euclidean distance measure is undertaken to assess the closeness. Hence, for a transition $(\boldsymbol{q}, a, \boldsymbol{q'})$, we can represent our reward function as follows:
\begin{equation}\label{reward_function}
\begin{split}
\mathcal{R}(\boldsymbol{q}, a, \boldsymbol{q'}) &= \text{sign}(d_{\boldsymbol{pq}}-d_{\boldsymbol{pq'}})\\
d_{\boldsymbol{ab}} &= ||\boldsymbol{a}-\boldsymbol{b}||_2
\end{split}
\end{equation}
where $\boldsymbol{p}$ is the target landmark position. Such binary reward is widely used in reinforcement learning and useful for tracking the progress. Even in the case of a continuous real-valued reward definition, it is recommended to perform reward clipping, where all the positive and negative outcomes are labelled as +1 and -1, respectively.

\subsection{Policy and value function}
Policy function outputs the optimal action-probabilities for a given state, whereas value function outputs the expected cumulative return for a given state and/or a given action. We adopt a stochastic policy to map states to actions. Previous RL-based localization approaches focused on implicit policy learning through training a value function approximator. We exploit a direct and explicit policy learning that comes under the category of policy-based RL, performing direct parametrisation of the policy function. In the proposed approach, a non-linear policy function approximator represented by a multi-layer perceptron (MLP) on top of a deep-CNN is used, embedding the high level feature learning from the raw state inside the policy learning. The parametrised policy function can be represented as follows:
\begin{equation}\label{policy_function}
\pi_{\theta}(\boldsymbol{q},a)=P(a|\mathcal{S}(\boldsymbol{q}),\theta)
\end{equation}
where $\theta$ represents the weights of the deep policy network.\par 

Direct policy search methods have a better convergence property while inducing a high variance problem. In actor-critic RL, an additional value function serves as a policy evaluator or critic to tackle the high variance problem. We use the state-value function approximator that tries to evaluate the policy, $\pi_\theta$, for the current policy parameters, $\theta$. We represent the value function approximator by another MLP stacked on top of the same CNN from the policy net, trying to approximate the state-value function to the true state-value (i.e., expected cumulative return for a state) for a given policy, as expressed as follows:
\begin{equation}\label{value_function}
V_{\omega}(\boldsymbol{q})\approx V^{\pi_\theta}(\mathcal{S}(\boldsymbol{q}))
\end{equation}
where $\omega$ refers to the network parameters of the value approximator network. Therefore, both the policy and value function share the parameters of a common CNN while having their own exclusive MLP parameters. Fig.~\ref{CNN}a presents the network architectures of the policy and value function in an actor-critic approach. \par

\subsection{Learning}
Using the aforementioned policy and value function approximator, actor-critic learning requires updating the parameters of both the policy (actor) and value (critic) networks. For a given state, actor-update aims at improving the policy to ensure a better cumulative return than the state-value inferred by the critic, whereas the critic aims at updating the value to approximate the cumulative return for the current policy. We perform the actor-critic RL in the widely used temporal differnce (TD) learning framework \cite{sutton1988learning}. We used the simplest linear TD(0) approach, where TD-target and TD-error are respectively calculated for a transition $(\boldsymbol{q}, a, \boldsymbol{q'})$ using the following equations: 
\begin{equation}\label{td_target}
\begin{split}
\tau(\boldsymbol{q}, a, \boldsymbol{q'}) &= \mathcal{R}(\boldsymbol{q}, a, \boldsymbol{q'}) + \gamma V_{\omega} (\boldsymbol{q'})\\
\varepsilon(\boldsymbol{q}, a, \boldsymbol{q'}) &= \tau(\boldsymbol{q}, a, \boldsymbol{q'}) - V_{\omega} (\boldsymbol{q})
\end{split}
\end{equation}
Thus, TD-target $\tau$ refers to the discounted cumulative return with a discount factor $\gamma$ using the current policy, and TD-error $\varepsilon$ refers to the advantage of the current policy over the critic-inferred state-value. Our goal is to approximate $\tau$ by the parametrised value function and update the policy towards the advantage. Hence, the cost functions for updating the value and policy parameters is stated as follows:
\begin{equation}
\begin{split}
& J_V(\omega) = \mathbb{E}_{(\boldsymbol{q}, a, \boldsymbol{q'})} \big[ \big(\tau(\boldsymbol{q}, a, \boldsymbol{q'}) - V_{\omega}(\boldsymbol{q}) \big)^2 \big]\\
& J_{\pi}(\theta) = \mathbb{E}_{(\boldsymbol{q}, a, \boldsymbol{q'})} \big[ - \varepsilon(\boldsymbol{q}, a, \boldsymbol{q'}) \log \pi_{\theta}(\boldsymbol{q},a) \big]
\end{split}
\end{equation}
In pure policy-based RL, expected return is used in the cost function of the policy network. The utilization of the advantage function formulated as TD-error in (\ref{td_target}) serves as a better cost function, enabling a low variance in the policy approximation.
\par


\section{Partial policy-based RL}
\label{sec4}
\begin{figure}[!ht]
\centering
\includegraphics[scale=1.0]{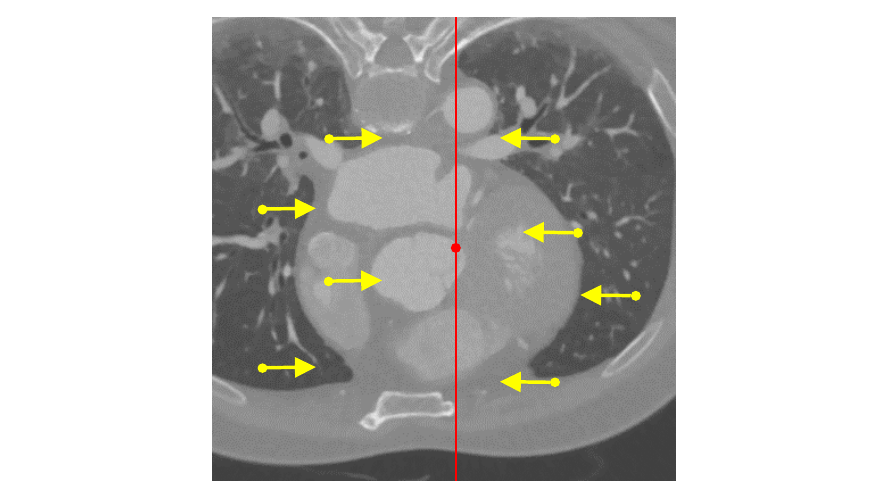}
\caption{{\bf Optimal policy problem in presence of alternative actions triggering positive feedback.} The depicted policy ensure rewards in all states by deciding between the actions along X (horizontal) axis only, except for the states lying on the red vertical line passing through the red dotted target landmark.}
\label{positive}
\end{figure}

RL agent learns the optimal behavior from episodes of experience gathered by interacting with the environment. In problems with large state/action space, successful task learning is challenging as it requires a huge number of trials, becoming a major drawback of RL. Despite the various methods of exploration, the issue still practically remains. Moreover, availability of multiple actions triggering a positive feedback for a state makes the optimal policy learning harder. In the case of 3D localization, we have six actions along X,Y and Z axes. Except for a negligible portion in the state space, a number of alternative actions can be found triggering a positive feedback for most of the states of a volume. Consequently, learning the decision between a pair of actions along one axis alone, can ensure positive feedbacks for almost the entire state space. Therefore, actions along other axes does not get significance to influence the policy update. The situation is illustrated in Fig.~\ref{positive}, where localization in a 2D slice is presented for simplicity. The agent using such biased policy is most likely to generate only the actions along X-axis, failing to learn the task.\par

For an efficient and effective learning of the optimal policy, we propose a partial policy-based learning approach. Instead of using one actor to update a large policy, we employ multiple micro-actors to learn the optimal behavior in partial sub-action spaces. Thus, the decomposition of the large problem into smaller problems enables an efficient, successful, and easier learning. Consequently, the proposed learning scheme can achieve the optimal policy within a fewer trials, compared to the conventional deep RLs. In this section, we first describe the dissection of the master policy to obtain partial policies, followed by the reconstruction of the original policy from the partial policies. Finally, we present the TD learning-framed actor-critic algorithm using the partial policies.\par

\subsection{Partial policy}
\begin{figure*}
\centering
\includegraphics[scale=0.9]{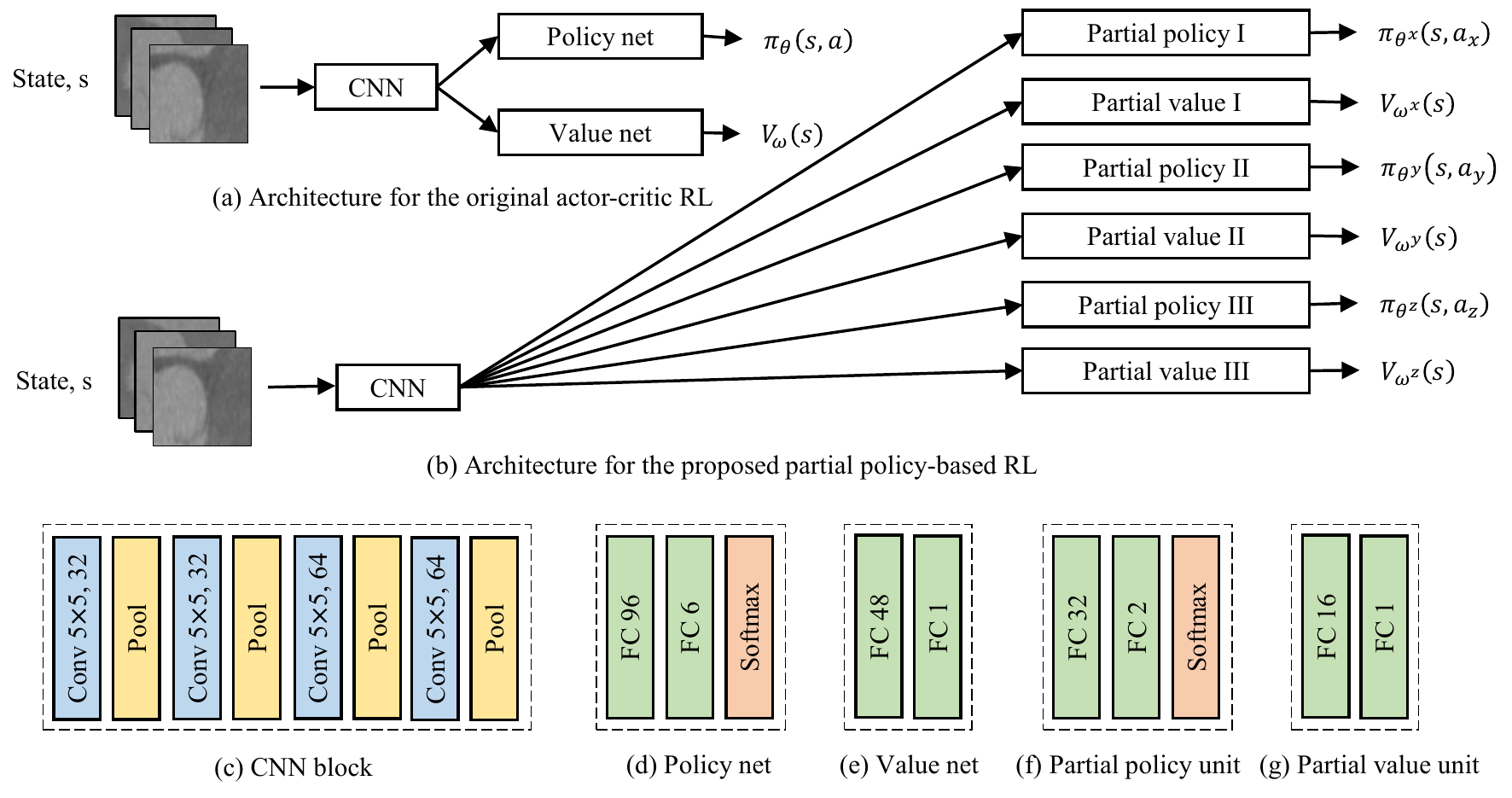}
\caption{{\bf Architectures of the original actor-critic RL and the proposed partial policy-based RL.} For a state, the policy net outputs probabilities for all possible actions, and the value net gives a single scalar value as the long-term return. FC stands for fully-connected layer.}
\label{CNN}
\end{figure*}

The objective of the partial policy is to obtain multiple simple policies on the projections of the actual action space, where the projected policies are able to reconstruct the policy on the original action space. We define smaller sub-action spaces i.e., partial action spaces, projecting the actual action space onto different Cartesian axes. Such dissection of the actual space also suggests multiple sub-agents corresponding to the partial action spaces, each of them trying to maximize the expected cumulative reward by taking optimal actions sampled from the corresponding sub-action space. Thus, we use multiple sub-agents learning smaller sub-tasks, instead of one agent with a large task. The actual action space in our 3D localization problem is decomposed into the following partial action spaces:
\begin{equation}\label{partial_action_space}
\begin{split}
\mathcal{A}_i=\{i^+,i^-\}, i \in \{x, y, z\}.
\end{split}
\end{equation}
Partial policy refers to the policy undertaken by the sub-agents to map state to actions in the corresponding axial domain. Thus, partial policies are projections of the original policy, defining the stochastic behavior for the partial action spaces. Therefore, we can define three partial policies with respect to the partial action spaces. Three independent MLPs sharing a common preceding CNN are used to represent the partial policies. To evaluate the partial policies, we also define three value function approximator networks stacked on top of the same CNN. Therefore, we have 6 MLPs preceded by a common CNN. For each sub-action space, there is a sub-actor and sub-critic available to update the corresponding partial policy and value function. The partial policy and value function approximators for our problem can be expressed as follows:
\begin{equation}\label{partial_policy_function}
\begin{split}
& \pi_{\theta^i}(\boldsymbol{q}, a_i)=P(a_i|S(\boldsymbol{q}),\theta^i), a_i \in \mathcal{A}_i, i \in \{x,y,z\} \\
& V_{\omega^i}(\boldsymbol{q})\approx V^{\pi_{\theta^i}}(S(\boldsymbol{q})), i \in \{x,y,z\}
\end{split}
\end{equation}  
where, $\theta^i$ and $\omega^i$ are the network parameters for the $i$-th partial policy and value approximators. Fig.~\ref{CNN} shows the overall network architecture.\par

Introducing the partial policies, the learning problem become easier and simpler in the shrunk action spaces. The goal of learning a partial policy is to provide the sub-agent at a state with the probability of the actions in the corresponding partial action space. The sub-agent only needs to choose between two actions to maximize the cumulative return. Therefore, the partial policy learning shares the same idea with the simplest binary classification problem. Attaining the optimal partial policy is easier, enabling a better convergence. The original definitions of state and reward function are used without any alteration.

\subsection{Reconstruction}
Partial policy ensuring a simpler learning is not adequate without the definition of actual policy reconstruction from the partial policies in order to employ it appropriately. sub-agents can decide the optimal partial actions from the learned partial policy. Deriving the actual action $\boldsymbol{a}_m$ combining the partial actions $a_i, i \in {x,y,z}$ is equivalent to sampling the action from the actual reconstructed policy. The actual action can be reconstructed as follows:
\begin{equation}\label{action_recon}
\begin{split}
\boldsymbol{a}_m=\sum_{i \in \{x,y,z\}}{a_i \zeta_i},\\
a_i \sim \pi_{\theta^i}(\boldsymbol{q}, a_i)
\end{split}
\end{equation} 
Here, $\zeta_i$ is the basis vector of the $i$-th axis. $\boldsymbol{a}_m$ is the actual action at position $\boldsymbol{q}$. \par
  
On the other hand, merging the partial policies to directly estimate the policy can be done by cascading the partial policies followed by normalization as expressed in the following:
\begin{equation}\label{policy_recon}
\begin{split}
\pi(\boldsymbol{q},:)= \frac{1}{3} \bigcup_{i \in \{x,y,z\}}\{\pi_{\theta^i}(\boldsymbol{q}, i^+), \pi_{\theta^i}(\boldsymbol{q}, i^-) \}
\end{split}
\end{equation} 
Using either approach of (\ref{action_recon}) and (\ref{policy_recon}) requires traversing all the three partial policy networks to estimate a single optimal behavior for a single state. Moreover, the critic evaluates a policy by quantifying the next state arrived after an action sampled from that policy. Utilizing the above reconstruction suggests a common next state obtained by a collaborative decision on the partial actions. This no longer holds the assumption of independent partial policy learning, again making the problem difficult.\par 

\begin{figure}
\centering
\begin{tikzpicture}[scale=0.7]
\begin{axis}[
    xlabel={X},
    ylabel={Z},
	zlabel={Y},    
    xmin=0, xmax=7,
    ymin=0, ymax=7,
    zmin=0, zmax=7,
    xtick={2,4,6},
    ytick={2,4,6},
	ztick={2,4,6},
    xmajorgrids=true,
    ymajorgrids=true,
    zmajorgrids=true,
    grid style=dashed,
]
 
\addplot3[
    ->,color=blue,
    mark=none,
    ]
    coordinates {
    	(2,2,2) (3,2,2)
    };
    \addlegendentry{$a_i \in \mathcal{A}_x \sim \pi_{\theta^x} ({\boldsymbol{q_i},a_i})$}
	\node[circle,fill,inner sep=1pt] at (axis cs:2,2,2) {};
	\node[] at (axis cs:1.8,2,1.4) {$q_0$};

\addplot3[
    ->,color=green,
    mark=none,
    ]
    coordinates {
    	(3,2,2) (3,2,3)
    };
    \addlegendentry{$a_i \in \mathcal{A}_y \sim \pi_{\theta^y} ({\boldsymbol{q_i},a_i})$}
	\node[circle,fill,inner sep=1pt] at (axis cs:3,2,2) {};
	\node[] at (axis cs:3.2,2,1.4) {$q_1$};

\addplot3[
    ->,color=red,
    mark=none,
    ]
    coordinates {
    	(3,2,3) (3,3,3)
    };
    \addlegendentry{$a_i \in \mathcal{A}_z \sim \pi_{\theta^z} ({\boldsymbol{q_i},a_i})$}
	\node[circle,fill,inner sep=1pt] at (axis cs:3,2,3) {};
	\node[] at (axis cs:2.6,2,3) {$q_2$};

\addplot3[
    ->,color=blue,
    mark=none,
    ]
    coordinates {
    	(3,3,3) (4,3,3)
    };
	\node[circle,fill,inner sep=1pt] at (axis cs:3,3,3) {};
	\node[] at (axis cs:2.6,3.4,3.2) {$q_3$};

\addplot3[
    ->,color=green,
    mark=none,
    ]
    coordinates {
    	(4,3,3) (4,3,4)
    };
	\node[circle,fill,inner sep=1pt] at (axis cs:4,3,3) {};
	\node[] at (axis cs:4.2,3,2.4) {$q_4$};

\addplot3[
    ->,color=red,
    mark=none,
    ]
    coordinates {
    	(4,3,4) (4,4,4)
    };
	\node[circle,fill,inner sep=1pt] at (axis cs:4,3,4) {};
	\node[] at (axis cs:3.6,3.4,4) {$q_5$};
	\node[circle,fill,inner sep=1pt] at (axis cs:4,4,4) {};
	\node[] at (axis cs:3.6,4.4,4.2) {$q_6$};
	
\end{axis}
\end{tikzpicture}
\caption{{\bf Sequential exploitation of the learned partial policies for localization.} Sub-agents are periodically exploited to make a number of step sequences, thus walking a Manhattan-like distance.}
\label{partial_in_action}
\end{figure}
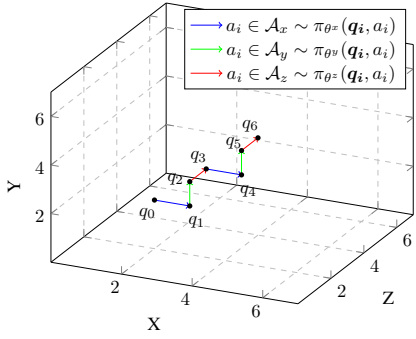

To ensure a greater efficiency, we propose a work-around to employ the partial policies to approximately represent the original policy. We suggest a periodic and sequential deployment of the partial policies, that can achieve the goal maintaining the efficiency. Thus, we perform a step-sequence instead of a single step. We define the $k$-th step-sequence $s_k$ as follows:
\begin{equation}\label{our_recon}
\begin{split}
& s_k=(a_{x,k}, a_{y,k}, a_{z,k})\\
& a_{x,k} \sim \pi_{\theta^x}(\boldsymbol{q}_t, a_{x,k})\\
& a_{y,k} \sim \pi_{\theta^y}(\boldsymbol{q}_{t+1}, a_{y,k})\\
&  a_{z,k} \sim \pi_{\theta^z}(\boldsymbol{q}_{t+2}, a_{z,k})\\
& \boldsymbol{q}_t = \mathcal{T}(\boldsymbol{q}_{t-1}, a_{z,k-1})\\
& \boldsymbol{q}_{t+1} = \mathcal{T}(\boldsymbol{q}_t, a_{x,k})\\
& \boldsymbol{q}_{t+2} = \mathcal{T}_(\boldsymbol{q}_{t+1}, a_{y,k})
\end{split}
\end{equation}
where $\boldsymbol{q}_t$ refers to the position at time step $t$, and $k = \lfloor \frac{t}{3} \rfloor$ is the order of step-sequence. We apply this sequence repeatedly, enabling a periodic selection of the partial policies. This also assures a balanced exploration in all sub-action spaces. The agent explicitly updates its position by taking a partial action from the defined sequence, contributing to the independent learning of the partial policies, because critic is able to give the feedback on the transitioned state solely reached by exploiting an individual partial policy. Only the responsible policy and value parameters are updated for a transition. The periodic application of the partial action sequence is depicted in Fig.~\ref{partial_in_action}. It is apparent that the actual agent is crossing a distance similar to Manhattan distance, periodically exploiting the sub-agents. The order of the partial actions in the unit sequence is not significant as long as they are repeated periodically in the overall action sequence. 

\subsection{Actor-critic RL for partial policy}
The periodic application of the partial actions sampled from the partial policy functions establishes the foundation of partial policy-based actor-critic learning. The same TD-framework is used maintaining the originality of the reward and transition, because those are not affected by the partial policy. Three independent micro-actors are responsible for updating the partial policy networks, each having a corresponding critic. \par 
At each step, one of the sub-agents are allowed to interact with the environment using the current parameters of the corresponding partial policy function, and utilize the critic to get directions to update the parameters. The critics also update the value based on the discounted return obtained by the current policy. Periodic deployment ensures the balance in learning all the partial policies and values. Algorithm ~\ref{partial_actor_critic_alg} presents the comprehensive actor-critic method for partial policy-based reinforcement learning. While original actor-critic method operates on step, the partial policy-based actor-critic method operates on unit sequence consisting of three partial steps. For each partial steps, parameter updates occur in the corresponding partial policy function as well as in the value function. Though we used batch methods to update the networks from experience replay, the algorithm presented here uses the incremental method for easier interpretation. In batch-methods, we first gather episodes of experience and store them in an experience replay memory, then sample mini-batches from the memory to perform a stochastic gradient descent. \par
\begin{algorithm}
\caption{Actor-critic RL using partial policy}
\label{partial_actor_critic_alg}
\begin{algorithmic}
\STATE Initialize $\boldsymbol{q}, \theta^x,\theta^y,\theta^z, \omega$
\FOR {each step sequence}
\FOR {$i \in \{x,y,z\}$}	
	\STATE Sample $a_i \sim \pi_{\theta^i}(\boldsymbol{q},a_i)$
	\STATE Next position, $\boldsymbol{q'}=\mathcal{T}(\boldsymbol{q}, a_i)$	
	\STATE Reward, $r= \mathcal{R}(\boldsymbol{q},a_i,\boldsymbol{q'})$
	\STATE TD-target, $\tau(\boldsymbol{q},a_i,\boldsymbol{q'})=r+\gamma V_{\omega^i} (\boldsymbol{q'})$
	\STATE TD-error, $\varepsilon(\boldsymbol{q},a_i,\boldsymbol{q'})=\tau(\boldsymbol{q},a_i,\boldsymbol{q'})-V_{\omega^i} (\boldsymbol{q})$
	\STATE $\theta^i \gets \theta^i + \alpha \nabla_{\theta^i} \varepsilon(\boldsymbol{q},a_i,\boldsymbol{q'}) \log \pi_{\theta^i} (\boldsymbol{q},a_i) $ 
	\STATE $\omega^i \gets \omega^i - \alpha \nabla_{\omega^i} (\tau(\boldsymbol{q},a_i,\boldsymbol{q'})-V_\omega (\boldsymbol{q}))^2$
	\STATE $\boldsymbol{q} \gets \boldsymbol{q'}$
\ENDFOR
\ENDFOR
\end{algorithmic}
\end{algorithm}

\begin{figure*}
\centering
\renewcommand{\arraystretch}{3.5}
\begin{tabular}{c c c}

\includegraphics[width=0.25\textwidth]{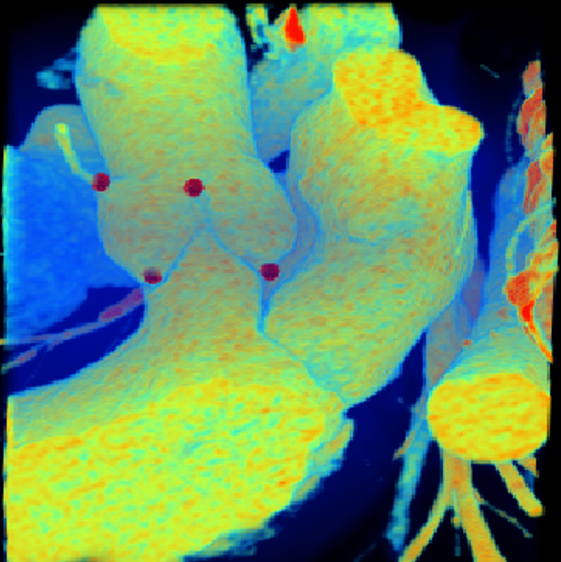}
&
\includegraphics[width=0.25\textwidth]{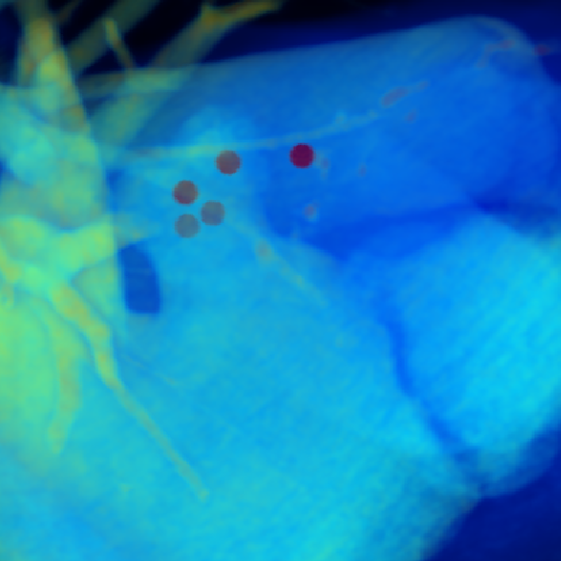}
&
\includegraphics[width=0.25\textwidth]{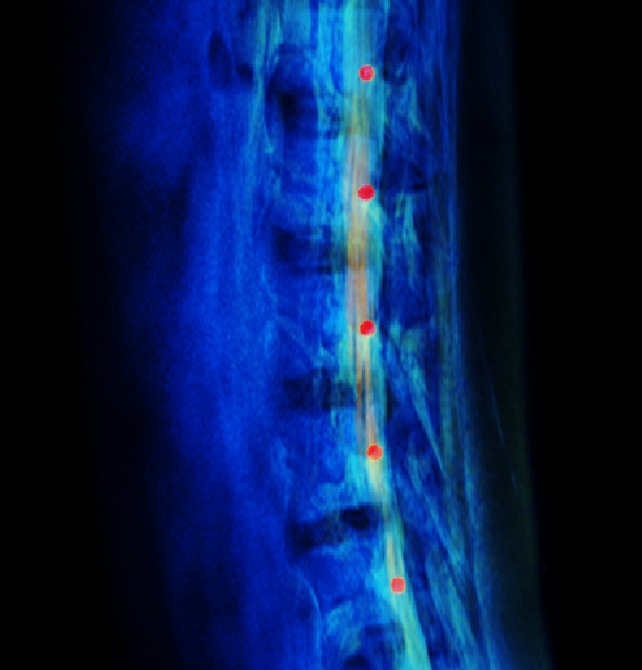}\\
\end{tabular}

\caption{{\bf Proposed partial policy based RL-localized landmarks.} (\textit{left}) Localized non-coronary and right coronary hinge points, the commisure point between them, and right ostium of the aortic valve in a CCTA volume are shown with other landmarks being occluded. (\textit{middle}) Localized LAA seed-points for different initial positions in a CT volume. (\textit{right}) Localized vertebra-centers in a spine MR volume.}
\label{fig_results}
\end{figure*}


\begin{table*}
\caption{Evaluation dataset and problem description.}
\centering
\renewcommand{\arraystretch}{1.5}
\setlength{\tabcolsep}{10pt}
\begin{tabular}{cccc}
\hline
{\bf } & {\bf Data} &  {\bf Voxel dimension} & {\bf Problem}\\
\hline
{\bf Dataset-A} & 71 coronary CT & 0.35 mm $\times$ 0.35 mm $\times$ 0.5 mm & Aortic valve landmarks\\
{\bf Dataset-B} & 150 cardiac CT & 0.45 mm $\times$ 0.45 mm $\times$ 0.5 mm & LAA seed-point\\
{\bf Dataset-C} & 20 spine MR (Public) & 0.58 mm $\times$ 0.58 mm $\times$ 1.5 mm & Vertebra centers\\
\hline
\end{tabular}
\label{dataset}
\end{table*}

\begin{table*}
\caption{Fourfold cross validation test results for localizing the aortic valve landmarks in CCTA volumes (Dataset-A) and left atrial appendage seed-point in CT volumes (Dataset-B). For Dataset-A, localization error is presented as the Euclidean distance from the ground truth position. The results for the TAVI and non-TAVI volumes are presented separately. For Dataset-B, the percentage of the localization failure is presented with respect to the total number of trials from different random position.}
\centering
\renewcommand{\arraystretch}{1.5}
\setlength{\tabcolsep}{10pt}
\begin{tabular}{c|c|cc|cc}
\cline{2-6}

\multicolumn{1}{c|}{ } & \multicolumn{1}{c|} {\multirow{2}{*}{\bf Landmark/method}} & \multicolumn{2}{c|} {\bf Partial Policy RL} & \multicolumn{2}{c} {\bf RL}\\

\multicolumn{1}{c|}{ } & \multicolumn{1}{c|}{ } & Mean $\pm$ SD &  Median &  Mean $\pm$ SD & Median\\
\hline
\multirow{3}{*}{Non-TAVI error (mm)} & Hinge points & 1.96 $\pm$ 0.98 & 1.72 & 2.18 $\pm$ 1.21 & 1.92\\

& Commissure points & 1.95 $\pm$ 0.93 & 1.69 & 2.16 $\pm$ 1.18 & 1.87\\

& Coronary ostia & 1.91 $\pm$ 0.95 & 1.65 & 2.13 $\pm$ 1.20 & 1.81\\
\hline
\multirow{3}{*}{TAVI error (mm)} & Hinge points & 2.08 $\pm$ 1.24 & 1.91 & 2.30 $\pm$ 1.27 & 2.11\\

& Commissure points & 2.02 $\pm$ 1.15 & 1.85 & 2.24 $\pm$ 1.22 & 2.06\\

& Coronary ostia & 1.94 $\pm$ 1.01 & 1.68 & 2.15 $\pm$ 1.32 & 1.94\\
\hline
\multirow{1}{*}{LAAO failure (\%)} & LAA & 7.21 $\pm$ 5.84 & 6.50 & 11.62 $\pm$ 7.71 & 9.22\\

\hline

\end{tabular}

\label{results_table}
\end{table*}

\section{Experiment}
\label{sec5}
We evaluate the proposed partial policy based RL method in three different problems in three datasets obtained from three different sites. Table~\ref{dataset} summarizes the evaluation datasets and the corresponding problems. The first dataset (Dataset-A) contains 71 contrast-enhanced coronary CT angiography (CCTA) volumes of 71 different patients. The corresponding problem is to localize the eight landmarks of the aortic valve (three hinge points, three commissure points, and two coronary ostia). Aortic valve (AV) landmark localization plays a vital role in preprocedural planning of transcatheter aortic valve implantation (TAVI) \cite{zheng}, which is an implant-based treatment method for severe aortic stenosis. Moreover, assessing the valve is a clinical routine during any cardiac CT interpretation. However, it is a time consuming task because the valve anatomy is not easily perceived in the conventional CT views. Among the 71 volumes, 31 volumes are preprocedural CT obtained from actual TAVI-patients. Accurate localization in TAVI volumes is challenging because valvular calcification can significantly affect the anatomy in unpredictable ways.\par

Using the second dataset (Dataset-B) consisting of 150 cardiac CT volumes, we localized the left atrial appendage (LAA) seed-point, which can facilitate an automatic segmentation of the appendage. LAA segmentation is helpful for physicians because it is a major site of thrombosis potentially responsible for inducing stroke-risk in non-valvular atrial fibrillation \cite{zoni2014epidemiology}. Related prior works proposed different segmentation approaches, however, within a manually marked bounding box (i.e., volume of interest) \cite{ROIFCNCRF}. The prior annotation of such bounding box enclosing LAA is a major obstacle of the approaches to become fully automatic. Therefore, localizing the aforementioned seed-point inside LAA can contribute to attaining an automatic segmentation method. Whereas the target points in the previous problem are specific, this problem suggests localizing any point inside the appendage. There is a large variation in appendage anatomy with an additional variation for different cardiac phase. The 150 volumes are obtained from 30 different patients in 5 different cardiac phases. \par

\begin{table}
\caption{Aortic valve landmark localization error comparison with different approaches.}
\centering
\renewcommand{\arraystretch}{1.5}
\setlength{\tabcolsep}{10pt}
\begin{tabular}{c|c}


\multicolumn{1}{c|}{\multirow{2}{*}{\bf Method}} & \multicolumn{1} {c} {\bf AV Localization error (mm)}\\
\multicolumn{1}{c|}{  } & \multicolumn{1} {c} {Mean $\pm$ SD}\\
\hline
Partial Policy RL & \multicolumn{1}{c}{1.98 $\pm$ 1.03}\\
Actor-critic & \multicolumn{1}{c}{2.19 $\pm$ 1.23}\\
DQN & \multicolumn{1}{c}{2.26 $\pm$ 1.35}\\
RTW \cite{cwplos} & \multicolumn{1}{c}{2.35 $\pm$ 1.48}\\
Inter-observer difference \cite{mustafaself} & \multicolumn{1}{c}{2.38 $\pm$ 1.56}\\
\hline

\end{tabular}
\label{comparison_table}
\end{table}
\begin{table}
\caption{Fourfold cross validation test results for localizing the vertebra centers in spine MR volumes (Dataset-C) using the partial policy-based RL.}
\centering
\renewcommand{\arraystretch}{1.5}
\setlength{\tabcolsep}{10pt}
\begin{tabular}{c|cc}


\multicolumn{1}{c|}{\multirow{2}{*}{\bf Vertebra}} & \multicolumn{2} {c} {\bf Localization error (mm)}\\
\multicolumn{1}{c|}{  } & \multicolumn{1} {c} {Mean $\pm$ SD} & \multicolumn{1} {c} {Median}\\

\hline
L1 & 2.79 $\pm$ 2.18 & 2.66\\

L2 & 2.61 $\pm$ 2.02 & 2.54\\

L3 & 2.58 $\pm$ 1.84 & 2.52\\

L4 & 2.86 $\pm$ 1.81 & 2.80 \\

L5 & 3.10 $\pm$ 2.08 & 3.05 \\
{\bf Overall} & {\bf 2.79 $\pm$ 1.98} & {\bf 2.71}\\ 
\hline
\cite{vertebra}'s result* & 2.87 $\pm$ 2.04 & 2.80 \\
\hline

\end{tabular}\\

*Same dataset but different train/test split
\label{vertebra_table}
\end{table}

The third dataset (Dataset-C) is a public dataset consists of 20 MR images of spine targeted at vertebra recognition for spine structure analysis \cite{vertebra}. This dataset is available at the SpineWeb online repository. We implemented our proposed method to localize the centers of the vertebra. We localized 5 lumbar vertebra (L1~L5). Among the 20 volumes, one has problematic ground truth (GT) annotation and one volume captured the head to shoulder region where the intended vertebra are not present. Therefore, we proceeded our experiment using 18 volumes.

For all the datasets, necessary ground truth positions of the target landmarks were obtained from the corresponding site, which were used to process the reward signals for the RL agent. For all the experiments, a common set-ups for RL is used. We implemented both the original actor-critic RL and the proposed partial policy-based RL for Dataset-A and B to obtain a comparative evaluation. To compare the proposed method with the widely-used DQN, we also implemented DQN for Dataset-A. For a fair comparison, we used identical parameters and hyper-parameters for all the methods. A window size of $m=50$ is used for the state, and the unit step size is set to 2 voxels. For each epoch, the agent was allowed to gather around 300 episodes of experience using its current policy, where each episode consists of 300 steps (or, 100 step-sequences in case of the partial policy). A replay memory of size $10^5$ is used to store the transitions. For the partial policy-based approach, three replay memories are maintained to keep track of the corresponding partial transitions. Sampling mini-batches from the experienced transitions, we perform stochastic gradient descend to update the policy and value network. The learning rate for updating both the value and policy was $\alpha = 10^{-4}$, and the discount factor $\gamma$ was set to 0.9. The CNN consists of 4 sets of convolutional, ReLU and max-pooling layer stacks (Fig.~\ref{CNN}). The final layer was flattened to obtain a non-spatial representation. 6 different MLPs were connected to the final flat layer of the CNN, representing the partial policy and value functions. For the original RL, only 2 MLPs are connected to represent a single set of policy and value functions. All the policy nets have a final softmax-gating to generate action-probabilities. Though $\epsilon$-greedy approach \cite{mnih2015human} is widely used for exploration in RL, Bayesian approach to allow the agent to define its own uncertainty has shown to perform better. Practically, the uncertainty is simulated by adding a dropout layer in the network \cite{gal2016dropout}. We gradually annealed the dropout keep probability from 0.1 to 0.7 over the epochs. To localize the vertebra-centers in Dataset-C using the proposed method, we use the same architecture and hyper-parameter settings.
\par

A four-fold cross validation is performed on patient-basis to evaluate the localization performance. For the first and third experiments, the localization error is calculated in terms of the Euclidean distance of the localized position from the corresponding ground truth. For the second experiment, such distance is not an appropriate measure because the assigned goal is to detect any point (seed) inside the left atrial appendage, and the annotated ground truth were also not specific. Therefore, the performance is measured using a binary comparison (i.e., whether the localized point was inside the appendage or not). During the test/validation session, the agent was not provided with any reward signal. For each test case, we conducted the localization process initiating the agent from different random positions inside the test volume, and presented the average localization result. Fig.~\ref{fig_results} depicts the qualitative localization results of the proposed partial policy-based RL. In Table~\ref{results_table}, we present the localization error (for Dataset-A) and localization failure percentage (for Dataset-B) of the proposed partial policy approach against the original actor-critic RL. Table~\ref{comparison_table} presents the average AV landmarks localization error comparison for different methods. Table~\ref{vertebra_table} presents the average localization error of the vertebra centers in spine MR images. The average computation time for localization is 1.2 seconds, as tested with a 3.60GHz single-core CPU and a GeForce GTX TITAN Xp GPU. Computation time for all the cases is same because an equal number of steps is performed in all cases.\par

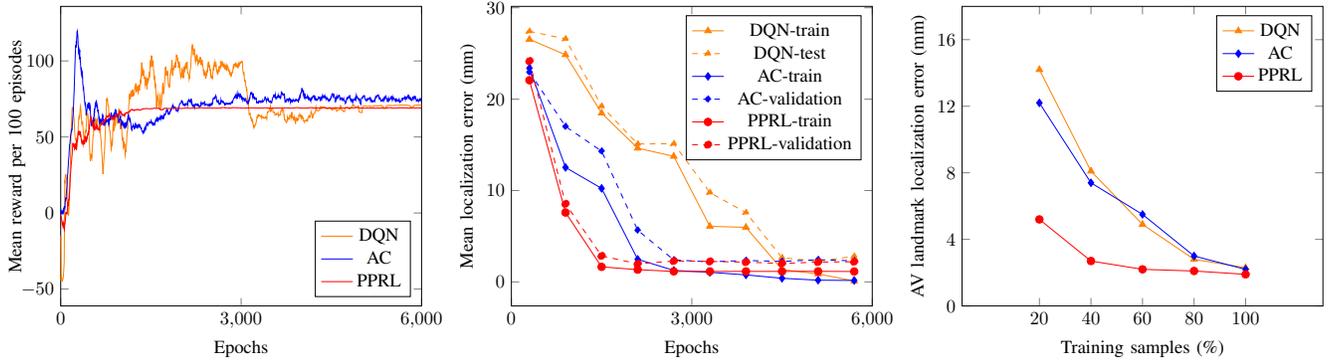
\begin{figure*}
\centering
\renewcommand{\arraystretch}{1.5}
\setlength{\tabcolsep}{1pt}
\begin{tabular}{c c c}
\begin{tikzpicture}[scale=0.7]
\begin{axis}[
     x label style={at={(axis description cs:0.5,0.0)},anchor=north},
    y label style={at={(axis description cs:0.1,.5)},anchor=south},     
    xlabel={Epochs},
    ylabel={Mean reward per 100 episodes},    
    xmin=0, xmax=6000,
    xtick={0,3000,6000},
    legend pos=south east,
]

\addplot [mark=none,color=orange] table [col sep=comma, mark=none] {rewards_dqn3.csv};
    \addlegendentry{DQN}
\addplot [mark=none,color=blue] table [col sep=comma, mark=none] {rewards_ac3.csv};
    \addlegendentry{AC}
\addplot [mark=none,color=red] table [col sep=comma, mark=none] {rewards_pprl3.csv};
    \addlegendentry{PPRL}

\end{axis}
\end{tikzpicture}
&
\begin{tikzpicture}[scale=0.7]
\begin{axis}[
     x label style={at={(axis description cs:0.5,0.0)},anchor=north},
    y label style={at={(axis description cs:0.1,.5)},anchor=south},     
    xlabel={Epochs},
    ylabel={Mean localization error (mm)},    
    xmin=0, xmax=6000,
    xtick={0,3000,6000},
    legend pos=north east,
]
\addplot[color=orange, mark=triangle*] table [col sep=comma] {dqn3train.csv};
    \addlegendentry{DQN-train}
    \addplot[color=orange, mark=triangle*, dashed] table [col sep=comma] {dqn3test.csv};
    \addlegendentry{DQN-test}
\addplot[color=blue, mark=diamond*] table [col sep=comma] {ac3train.csv};
    \addlegendentry{AC-train}
\addplot[color=blue, mark=diamond*, dashed] table [col sep=comma] {ac3test.csv};
    \addlegendentry{AC-validation}
\addplot[color=red, mark=otimes*] table [col sep=comma] {pprl3train.csv};
    \addlegendentry{PPRL-train}
\addplot[color=red, mark=otimes*, dashed] table [col sep=comma] {pprl3test.csv};
    \addlegendentry{PPRL-validation}
\end{axis}
\end{tikzpicture}
&
\begin{tikzpicture}[scale=0.7]
\begin{axis}[
	x label style={at={(axis description cs:0.5,0.0)},anchor=north},
    y label style={at={(axis description cs:0.1,.5)},anchor=south},  
    xlabel={Training samples (\%)},
    ylabel={AV landmark localization error (mm)},    
    xmin=-10, xmax=130,
    ymin=0, ymax=18,
    xtick={ 20,40,60,80,100},
    ytick={0,4,8,12,16},
    legend pos=north east,
]

\addplot[color=orange, mark=triangle*]
coordinates{(20, 14.2) (40, 8.1) (60,4.9) (80,2.8) (100, 2.3)};
	\label{plot_three}    
    \addlegendentry{DQN}
    
\addplot[color=blue, mark=diamond*]
coordinates{(20, 12.2) (40, 7.4) (60,5.5) (80,3.0) (100, 2.2)};
	\label{plot_two}    
    \addlegendentry{AC}
    \addplot[color=red, mark=otimes*] 
coordinates{(20, 5.2) (40, 2.7) (60,2.2) (80,2.1) (100, 1.9)};
	\label{plot_one}
    \addlegendentry{PPRL}

\end{axis}

\end{tikzpicture}

\end{tabular}
\caption{{\bf Learning curves of the proposed partial policy-based RL.} (\textit{left} and \textit{middle}) Average reward and localization error over different epochs, for AV landmark localization. Reward plot is smoothed out for better visualization. (\textit{right}) Localization performance with respect to training data size. Training set size is gradually increased while validating with a common test set. The proposed method could achieve almost the maximum accuracy with a fewer training examples.}
\label{fig_learning_curves}
\end{figure*}
The proposed method showed an average error of 1.98 $\pm$ 1.03 mm localizing the AV landmarks in CCTA volume, whereas Elattar et al. \cite{mustafaself}'s error for localizing the hinge points and ostia in CTA was 2.65 $\pm$ 1.57 mm, and Zheng et al. \cite{zheng}'s error for localizing all the landmarks in C-arm CT was $2.11 \pm 1.34$ mm. The inter-observer difference in CTA was $2.38 \pm 1.56$ mm, as reported in \cite{mustafaself}. The proposed method also showed improvement comparing to the random tree walk method in our previous work using the same dataset, where the average localization error was 2.35 $\pm$ 1.48 mm \cite{cwplos}. On the other hand, RTW takes only a few milliseconds to localize a landmark because of its simple feature computation. However, CNN-based feature computation in the currently proposed method is more useful and the current localization time of 1.2 seconds can still be considered efficient and allowable for such an improved accuracy. In our previous work, we also introduced a colonial walk method utilizing multiple walks from multiple initial positions and  choosing the final walk by minimum walk variance. Such extension can also be performed with the current RL agent walk to improve the accuracy even more, which we keep as our future work. As for the center locations of the vertebra in spine MR volumes, the proposed method showed an average error of 2.79 $\pm$ 1.98 mm, which is as good as \cite{vertebra}'s result.\par

\begin{figure}
\centering
\includegraphics[scale=1.0]{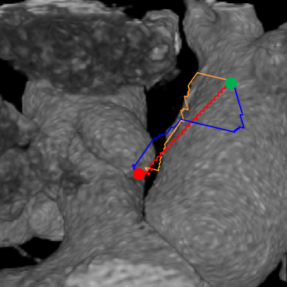}
\caption{{\bf Search paths of different RL agents localizing a hinge point.} The transitions of the PPRL, acotr-critic, and DQN agents is indicated by red, blue, and orange lines, respectively. Green and red dots indicate the initial position and target landmark.}
\label{search_path}
\end{figure}

The proposed partial-policy based method exhibited noteworthy improvement over the original actor-critic and the DQN approach. The average localization error is improved with a significant reduction of error variance. Learning the partial policies facilitated an improved localization with simpler decision process. At each step sequence, the sub-agents solve three binary decision problems. Consequently, the proposed method exhibited a noteworthy improvement comparing to directly learning the original policy. We also performed an additional experiment where the agent is trained only on the 40 non-TAVI volumes and attempts to localize the landmarks in the TAVI volumes. Thus, we could observe the agent behavior in a volume with valvular calcification, without providing any prior knowledge about calcified valves. Table~\ref{results_table2} presents the average localization results for the proposed method and the conventional actor-critic RL. Partial policy could cope with the variation due to calcification significantly better than the conventional RL because of simpler decision space.

\begin{table}[!ht]
\caption{Aortic valve landmark localization results in calcified TAVI volumes by an agent trained on only non-TAVI volumes.}
\centering
\renewcommand{\arraystretch}{1.5}
\setlength{\tabcolsep}{10pt}
\begin{tabular}{c|c|c}
\hline
\multicolumn{1}{c|} {\multirow{2}{*}{\bf Localization error (mm)}} & \multicolumn{1}{c|} {\bf Partial Policy RL} & \multicolumn{1}{c} {\bf RL}\\

\multicolumn{1}{c|}{ } & Mean $\pm$ SD &  Mean $\pm$ SD \\
\hline
 Hinge points & 2.52 $\pm$ 1.78 & 3.38 $\pm$ 2.19 \\

Commissure points & 2.35 $\pm$ 1.74 & 3.16 $\pm$ 2.17\\

Coronary ostia & 2.26 $\pm$ 1.68 & 2.97 $\pm$ 2.02 \\

\hline

\end{tabular}

\label{results_table2}
\end{table}

To compare the learning process and learned trajectories, we plot the average reward and localization error over the epoch from the learning process of the proposed method and the conventional ones, and illustrate optimal search paths in Fig.~\ref{fig_learning_curves}. A remarkable improvement is observed in the learnability of the proposed partial policy approach. It enabled a better and faster convergence. It converges within about half the epochs required by the conventional RLs (i.e., actor-critic and DQN), thus improving the slow learning problem in RL. Within a few-trials, the sub-agents could reach an optimal behavior, as depicted in the error plots. The search paths of the sub-agents also exhibit more confident transitions compared to the paths undertaken by the conventional agents (Fig.~\ref{search_path}). \par

Preparing training data with ground truth acquisition is a difficult task in medical image processing. Therefore, it is advantageous to have a model that can give a satisfactory performance with knowledge of a fewer training data. Apart from the standard train/test splits, we randomly sampled about 20\% of the dataset to be the test data (for Dataset-A). From the rest of the dataset, we gradually sampled 20\%, 40\%, 60\%, 80\%, and 100\% to obtain five training subsets, where the last subset (100\% training samples) refers to the whole training set. The test set is validated by the models trained on those subsets. For comparison, we used an identical split for the proposed PPRL, as well as the actor-critic and DQN approach. Fig.~\ref{fig_learning_curves}(\textit{right}) presents our observation, where the proposed method could achieve very close to the maximum accuracy (achieved with 100\% training samples) with a notably fewer training exmaples. Even with 20\% of the training data, it could provide an average error of 3.8 mm, which is comparable to the  10 mm error of DQN and actor-critic. Thus, the partial-policy based RL can potentially be useful in medical applications.\par 


\section{Conclusion}
\label{sec6}
Anatomical Landmark localization provides significant prior information for different applications in medical image processing. Our work presented a robust localization method formulated as reinforcement learning. For an efficient and successful learning with actor-critic method, we introduced a partial policy-based learning where multiple easier policies are learned on the sub-action spaces defined as projections of the original action space. Employing multiple sub-agents interacting with the environment, corresponding micro-actors and micro-critics  independently update the deep partial policy and value networks, enabled a faster and better convergence. The experiment with aortic valve landmarks localization and left atrial appendage seed localization in 3D CT images, and vertebra localization in 3d spine MR images showed robust and improved performance, compared to the conventional actor-critic and widely used deep Q-learning approach. The proposed partial policy based approach required significantly fewer number of trials and fewer training data to achieve the optimal behavior, improving the slow learning problem of RL. The proposed method provides an efficient and potentially useful solution for localization, requiring an average localization time of 1.2 seconds.


%

\section*{Acknowledgment}

This research was supported by Basic Science Research Program through the National Research Foundation of Korea (NRF), funded by the Ministry of Education, Science, Technology (No. 2017R1A2B4004503).

\ifCLASSOPTIONcaptionsoff
  \newpage
\fi



%

\bibliographystyle{IEEEtran}
\bibliography{mybib}

%





\end{document}